\documentclass{article}
\usepackage{ijcai25}
\usepackage{times}
\usepackage{soul}
\usepackage{url}
\usepackage[hidelinks]{hyperref}
\usepackage[utf8]{inputenc}
\usepackage[small]{caption}
\usepackage{graphicx}
\usepackage{amsmath}
\usepackage{amsthm}
\usepackage{booktabs}
\usepackage{algorithm}
\usepackage{algorithmic}
\usepackage[switch]{lineno}
\usepackage{threeparttable}
\usepackage{tabularx}
\usepackage{caption}
\usepackage{colortbl}
\usepackage[table]{xcolor}
\usepackage{rotating}
\usepackage{multirow} 
\usepackage{makecell}
\usepackage{bbding}
\usepackage{subfigure}
\usepackage{amsfonts}
\usepackage{amsmath}
\usepackage{bm}
\usepackage{amssymb}

\definecolor{gray50}{gray}{0.5}
\definecolor{python1}{rgb}{0.11, 0.451, 0.702}
\definecolor{python2}{rgb}{0.988, 0.486, 0.047}

\newtheorem{theorem}{Theorem}

\title{SDDiff: Boost Radar Perception via Spatial-Doppler Diffusion}

\iftrue
\author{
Shengpeng Wang$^1$
\and
Xin Luo$^1$\and
Yulong Xie$^{1}$\And
Wei Wang$^{2}$ \thanks{Corresponding author}\\
\affiliations
$^1$Huazhong University of Science and Technology\\
$^2$Wuhan University\\
\emails
\{wsp666, l\_xin, yulong\_xie\}@hust.edu.cn, wangw@whu.edu.cn
}
\fi

\begin{document}

\maketitle

\begin{abstract}
    Point cloud extraction~(PCE) and ego velocity estimation~(EVE) are key capabilities gaining attention in 3D radar perception. However, existing work typically treats these two tasks independently, which may neglect the interplay between radar's spatial and Doppler domain features, potentially introducing additional bias. In this paper, we observe an underlying correlation between 3D points and ego velocity, which offers reciprocal benefits for PCE and EVE. To fully unlock such inspiring potential, we take the first step to design a \textbf{S}patial-\textbf{D}oppler \textbf{Diff}usion~(SDDiff) model for simultaneously dense PCE and accurate EVE. To seamlessly tailor it to radar perception, SDDiff improves the conventional latent diffusion process in three major aspects. First, we introduce a representation that embodies both spatial occupancy and Doppler features. Second, we design a directional diffusion with radar priors to streamline the sampling. Third, we propose Iterative Doppler Refinement to enhance the model’s adaptability to density variations and ghosting effects. Extensive evaluations show that SDDiff significantly outperforms state-of-the-art baselines by achieving 59\% higher in EVE accuracy, $4\times$ greater in valid generation density while boosting PCE effectiveness and reliability.
\end{abstract}

\section{Introduction}
\begin{figure}[t]
  \centering
   \includegraphics[width=\linewidth]{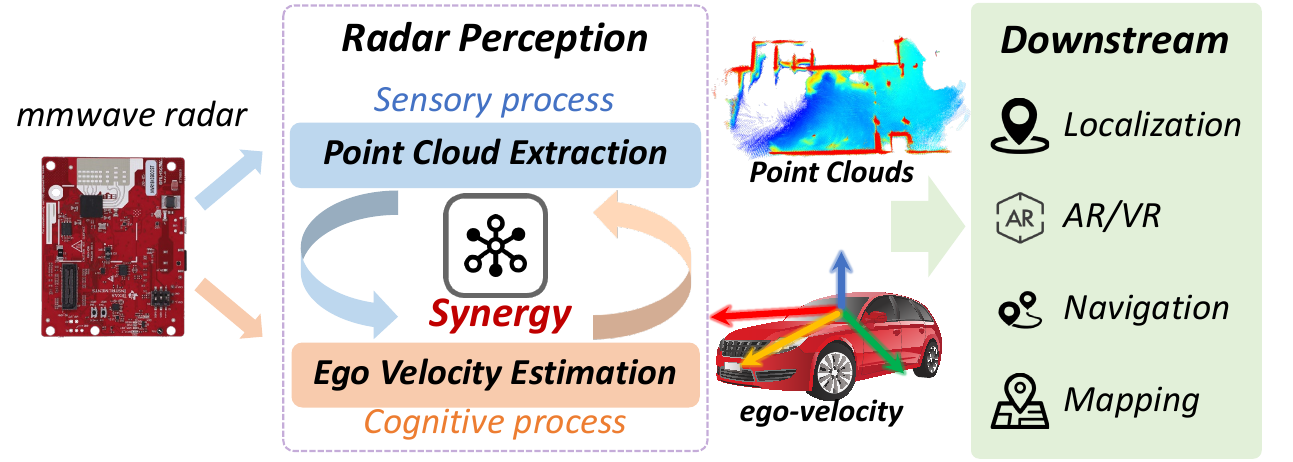}
   \caption{Illustration of our work's objective: simultaneously extracting point clouds and estimating ego velocity to enhance radar perception.}
   \label{fig:map}
\end{figure}

Millimeter-wave radar for all-weather perception is increasingly attracting widespread attention in robotics, computer vision, augmented reality, and autonomous driving~\cite{harlow2024new,zhang2022sar,xu2021followupar,adolfsson2021cfear}. Point cloud extraction~(PCE) and ego velocity estimation~(EVE) are crucial pillars of radar perception. PCE acts as a \textit{low-level sensory process}, extracting fundamental object information from reflected radar signals, including position, reflectivity, and Doppler velocity. Conversely, EVE serves as a \textit{high-level cognitive process}, leveraging elemental point clouds to infer the radar’s ego velocity. As Fig. \ref{fig:map} shows, comprehensive PCE and accurate EVE form a solid perceptual foundation for downstream tasks, including object detection, simultaneous localization and mapping~(SLAM), path planning, and autonomous navigation.

\begin{figure*}[h]
	\centering
	\subfigure[]{
		\label{level.sub.1}
		\includegraphics[width=0.24\linewidth]{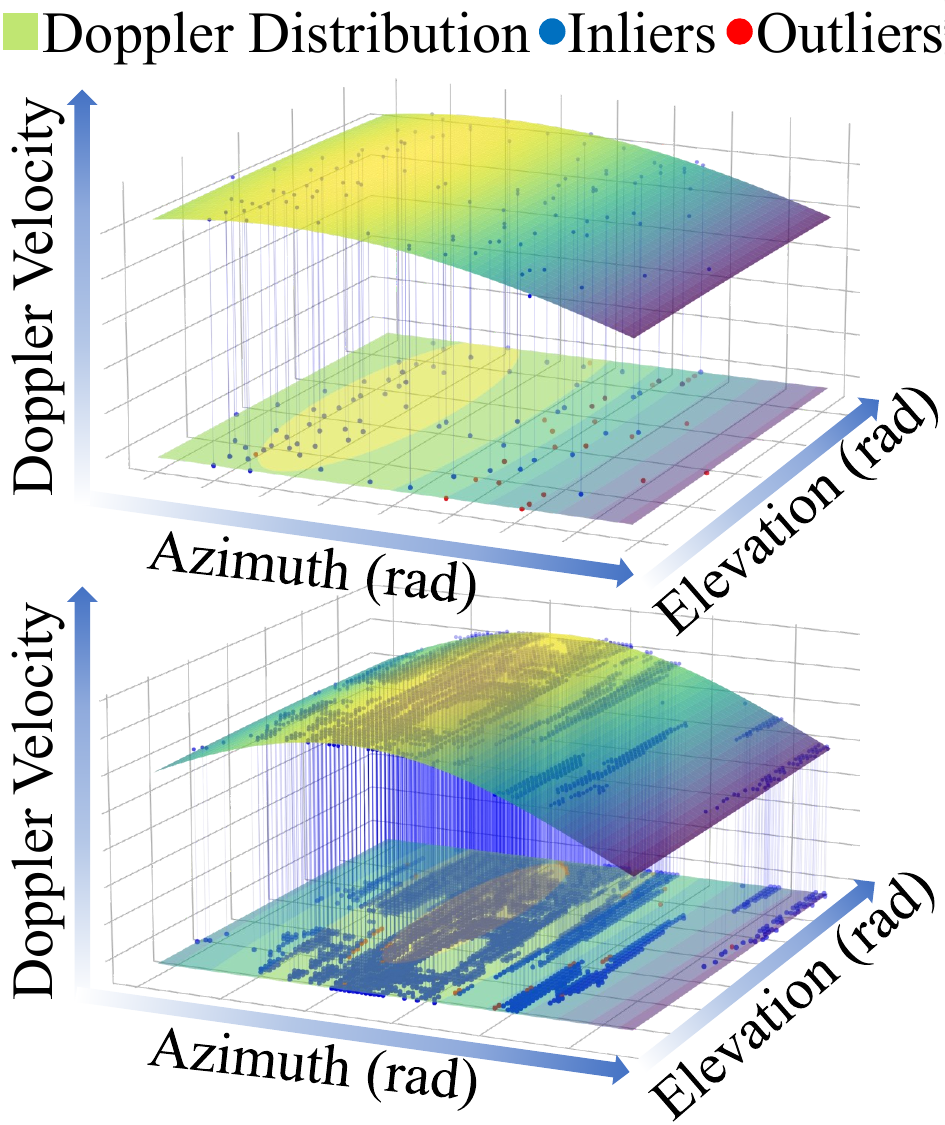}}
	\subfigure[]{
		\label{level.sub.2}
		\includegraphics[width=0.24\linewidth]{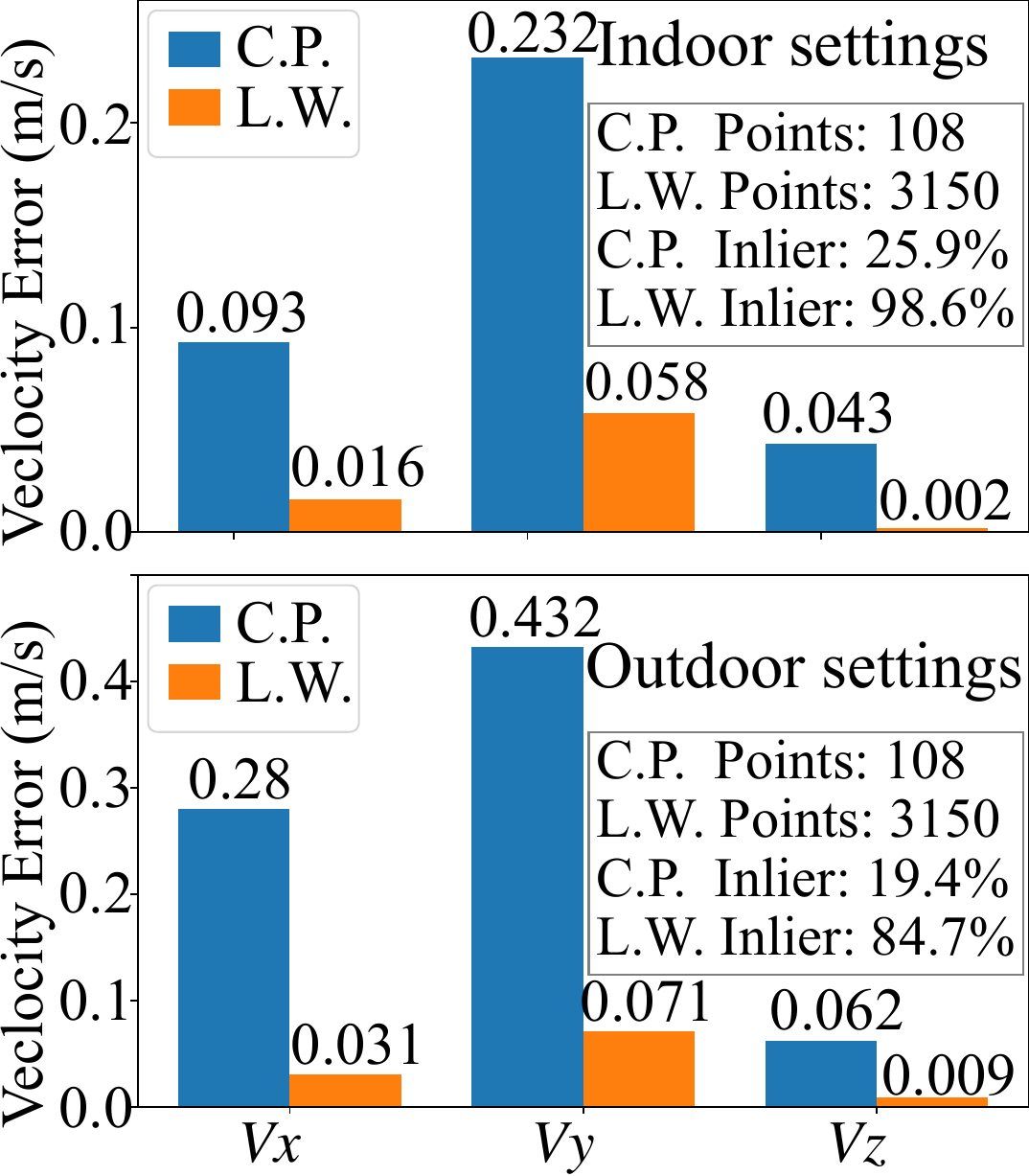}}
        \subfigure[]{
		\label{level.sub.3}
		\includegraphics[width=0.48\linewidth]{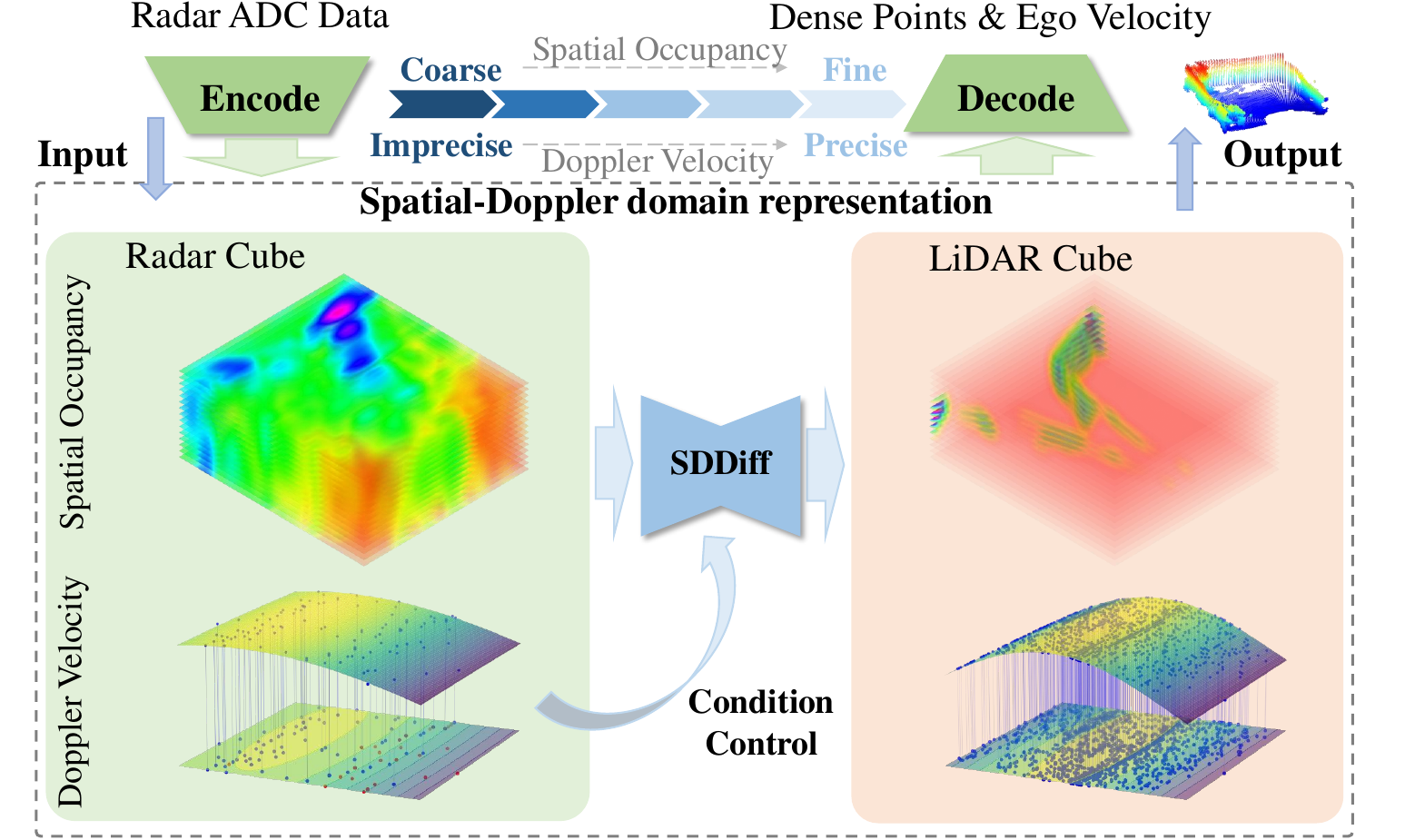}}
	\caption{(a)~The \textcolor{python1}{upper} figure presents Doppler velocity distribution derived from sparse points after onboard CFAR post-processing, while the \textcolor{python2}{lower} figure displays points warped by LiDAR. (b)~Given CFAR-processed~(C.P.) points or LiDAR-warped~(L.W.) points, velocity estimation errors are observed in both indoor and outdoor settings. (c)~The brief illustration of SDDiff, where PCE and EVE are jointly refined through the SDDR Purification Process, sculpting coarse representations into fine ones.}
	\label{fig:motivation}
\end{figure*}
Existing studies have started to focus on PCE~\cite{prabhakara2023high,10592769} and EVE~\cite{pang2024radarmoseve} separately, yielding promising results. These tasks are typically treated independently, with PCE relying solely on signal intensity and EVE anchored in sparse, chaotic points processed by onboard systems. However, addressing these tasks in isolation may overlook the interplay between radar's spatial and Doppler domain features, potentially introducing additional bias. Specifically, on the one hand, non-object regions may display high reflection intensity due to the notorious ``multi-path effect"~\cite{yataka2024sira}. Consequently, relying solely on signal intensity may mislead PCE, impeding the balance between detection density and clutter suppression. On the other hand, sparse, noisy points with poor vertical angle resolution severely degrade EVE performance.

One of our key observations is that there is a synergy between PCE and EVE, which can reciprocally evoke potential gains from each other. We conduct a feasibility study using the Coloradar Dataset~\cite{kramer2022coloradar}, as shown in Fig.~\ref{level.sub.1}. The orientation and Doppler velocity of 3D point clouds jointly define a surface, with its precise parameters determined by the radar's ego velocity. Given various points, we estimate these parameters by Random Sample Consensus~(RANSAC)~\cite{fischler1981random}. As shown in Fig.~\ref{level.sub.2}, sparse, noisy points after onboard CFAR post-processing lead to an awkward inlier rate~(about 25.9\% at a threshold of 0.08). Points warped by dense LiDAR point clouds show over 5$\times$ improvement in EVE, with the inlier rate reaching 98.6\% indoors. This demonstrates that the synergy offers reciprocal benefits: accurate ego velocity helps filter or refine points aligned with Doppler velocity. In turn, robust and dense PCE can further enhance EVE accuracy.

To fully unlock such inspiring reciprocal potential, we introduce a Spatial-Doppler Domain Representation~(SDDR) for 3D radar perception. As shown in Fig.~\ref{level.sub.3}, the radar analog-to-digital converter~(ADC) signal is encoded into a 3D tensor that embodies both spatial occupancy and Doppler velocity. PCE and EVE are jointly refined through SDDR Purification Process, sculpting coarse, ghost-prone representations into fine, uncontaminated ones.

Inspired by remarkable generative capabilities of diffusion models~\cite{ho2020denoising}, we design a \textbf{S}patial-\textbf{D}oppler \textbf{Diff}usion~(SDDiff) model to generate an enhanced SDDR by effectively aligning spatial and Doppler features. Specifically, we propose two key designs that allow SDDiff to seamlessly tailor to 3D radar perception. On the one hand, naive diffusion model aims to transform multi-modality in latent space, requiring a plethora of sampling steps. This high demand is attributed to sampling from a standard Gaussian distribution, which is well-suited for image synthesis. However, it introduces sampling wastage and ambiguous outcomes for the SDDR purification task. To address this, we design a directional diffusion with radar priors, transforming consistent modality in SDDR. This significantly reduces the sampling steps and facilitates finer spatial occupation. On the other hand, we incorporate iteratively refined Doppler velocity profiles into the model as conditions and apply a physical constraint to bridge PCE and EVE based on the synergy. This enables the model to learn adaptive feature representations that are resilient to density variations and ghost effects.

We summarize the contributions below:
\begin{itemize}
     \item We propose a novel Spatial-Doppler Diffusion model to sculpt coarse representations into fine ones for 3D radar perception. To the best of our knowledge, this is the first attempt to simultaneously attain dense point clouds and accurate ego velocity from Spatial-Doppler Domain Representation.
    \item We design a directional Spatial-Doppler diffusion with radar priors to streamline the sampling. This deeply reduces sampling steps and mitigates ambiguous outcomes.
    \item We propose Iterative Doppler Refinement, leveraging the Doppler-consistency of refined spatial occupancy to enhance the model’s adaptability to density variations and ghosting effects.
    \item  Experimental results demonstrate that our method outperforms the previous state of the arts and achieves reciprocal benefits across PCE and EVE. Additionally, we will make our self-collected dataset publicly available to the research community. This dataset facilitates further advancements in 3D radar perception.
\end{itemize}

\section{Related Work}

\subsection{Traditional Model for Radar Detection}
In early radar detection, low-pass filtered intermediate frequency~(IF) signals from millimeter-wave radar were processed via FFT, angle estimation~\cite{schmidt1986multiple,li2003robust,roy1989esprit}, and Constant-False-Alarm-Rate~(CFAR)~\cite{nitzberg1972constant}, resulting in sparse, artifact-laden point clouds. Such sparse points are insufficient for downstream tasks such as state estimation, mapping, and navigation. To generate denser radar points, some research~\cite{lai2024enabling,qian20203d} leverages radar motion to construct virtual antenna arrays, similar to synthetic aperture radar~(SAR) imaging~\cite{ausherman1984developments}, to enhance angular resolution. However, these methods depend on precise motion estimation or predetermined trajectory, hindering their translation into real-world implementation. Other studies~\cite{cen2019radar,cen2018precise} concentrate on eliminating clutter points caused by multipath effects to enhance the quality of radar point clouds. However, they are limited to high-resolution mechanical scanning radars and unsuitable for commercial off-the-shelf radar.

\subsection{Generative Model for Radar Detection}
Recently, numerous studies have focused on generating denser radar points through cross-modal supervision from high-resolution sensors, such as LiDAR or depth camera. Guan et al.~\cite{guan2020through} leverages a conditional Generative Adversarial Network~(cGAN)~\cite{goodfellow2014generative,sun20213drimr} to achieve object-level imaging and recover the high-frequency shape of the specific object. To materialize scene-level imaging, milliMap~\cite{lu2020see} and RadarHD~\cite{prabhakara2023high} employ a frame-stacking strategy, combining multiple CFAR-filtered Cartesian map patches or planar Range-Azimuth~(RA) maps as U-Net~\cite{ronneberger2015u} inputs, with maps projected by LiDAR points for supervision. Similarly, Zhang et al.~\cite{10592769} apply a diffusion model~\cite{ho2020denoising} to restore radar RA maps under LiDAR supervision, subsequently generating radar points in bird’s-eye view~(BEV). However, these intensity-only methods overlook the significant contributions of Doppler features to the spatial point cloud's contextual understanding. To enhance 3D point clouds, Luan et al.~\cite{10611026} encode CFAR-processed 3D sparse point clouds into BEV images for diffusion model. Unlike the direct approaches, other studies~\cite{cheng2022novel,fan2024enhancing} predict Range-Doppler~(RD) maps, and then use conventional angle estimation to derive 3D points. Nevertheless, they still resort to traditional angle estimation.
\subsection{Ego Velocity Estimation for Radar}

Some methods estimate ego velocity by solving a transformation between two consecutive exteroceptive measurements. General registration techniques, such as ICP~\cite{besl1992method}, NDT~\cite{biber2003normal}, and their variants~\cite{censi2008icp,segal2009generalized}, are often applied for this. However, these sensor-agnostic approaches struggle with noisy radar point clouds, which provide poor point-to-point correspondence. While~\cite{cen2019radar} introduces a keypoint extraction and data association scheme, it is unsuitable for low-resolution commercial radars. More methods leverage Doppler velocities for 2D ego velocity estimation, typically relying on RANSAC~\cite{kellner2013instantaneous,kellner2014instantaneous} or end-to-end networks like RadarEVE~\cite{pang2024radarmoseve}. However, these approaches are built on sparse and chaotic points processed by the radar's onboard system. In open environments with few points, such methods suffer significant performance degradation or even complete failure. To address these challenges, we propose SDDiff, the first method, to the best of our knowledge, that enables both dense point cloud extraction (PCE) and accurate 3D ego velocity estimation (EVE) directly from the raw ADC data of a single-chip radar.

\section{Spatial-Doppler Diffusion Model}
In this section, we present the Directional Spatial-Doppler Diffusion Model with radar priors in Sec.~\ref{DDwrp} and Iterative Doppler Refinement in Sec.~\ref{IDR}, followed by an overview of SDDiff model as illustrated in Sec.~\ref{OOS}.

\subsection{Directional diffusion with radar priors}\label{DDwrp}

\begin{figure}[h]
  \centering
   \includegraphics[width=.8\linewidth]{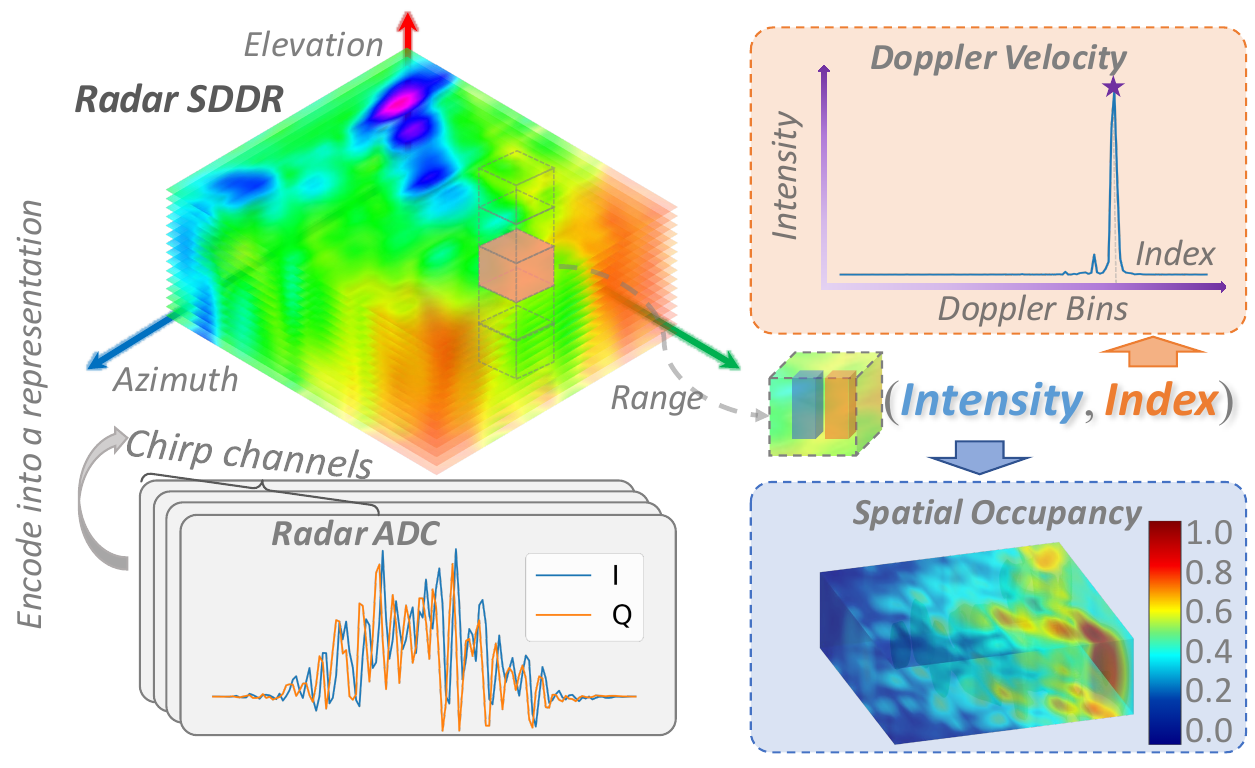}
   \caption{Diagram of Spatial-Doppler Domain Representation.}
   \label{fig:sddr}
\end{figure}

\noindent \textbf{Spatial-Doppler Domain Representation}
We focus on radar's information-rich ADC data rather than the sparse points from onboard systems. To simultaneously achieve dense PCE and accurate EVE, we introduce a Spatial-Doppler Domain Representation~(SDDR) designed around two key principles: 1) Capturing both spatial occupancy and Doppler velocity with minimal information loss. 2) Eliminating redundant data to reduce computational cost and memory usage when feasible. Specifically, we transform raw radar data into a 4D tensor $\boldsymbol{C'} \in \mathbb{R}^{R \times A \times E \times D}$ via fast Fourier transform (FFT), where the dimensions correspond to range, azimuth, elevation, and Doppler. Higher values in the tensor typically indicate a higher likelihood of a point's existence. For a spatial position $\bm{s}_{k,i,j}=(r_k,a_i,e_j)$, the Doppler velocity $v^r_{k,i,j}$ is uniquely determined. Empirically, the maximum value along the Doppler axis is typically at least 6$\times$ greater than the second when a point exists at $\bm{s}_{k,i,j}$. Using this property, we extract the indices of peak values along the Doppler axis to determine Doppler velocities as shown in Fig.~\ref{fig:sddr}. Subsequently, intensity $\bm{u}$ and Doppler velocity $\bm{v}$ are concatenated into a polar SDDR $\bm{C}=[\bm{u};\bm{v}] \in \mathbb{R}^{R\times A \times E \times 2}$, where intensity represents the spatial occupancy and the index corresponds to the Doppler velocity. This can significantly reduce the data volume and computational overhead.

Similarly, LiDAR's spatial occupancy can be represented by wrapping the points into a polar format consistent with the radar configuration. Refining PCE and EVE is defined as the SDDR Purification Process, transforming coarse, ghost-prone representations into fine, uncontaminated ones. The conventional diffusion process typically transforms the target distribution into a standard Gaussian distribution over long steps, and subsequently iteratively samples a new target starting from the Gaussian noise given conditional embeddings. However, it requires a plethora of sampling steps and produces ambiguous outcomes for the SDDR purification task. In response, we define a directional diffusion process, starting from an initial state with a homogeneous radar representation. The forward and reverse processes are comprehensively presented and analyzed as follows.

\begin{figure*}[h]
	\centering 
	\includegraphics[width=.98\linewidth]{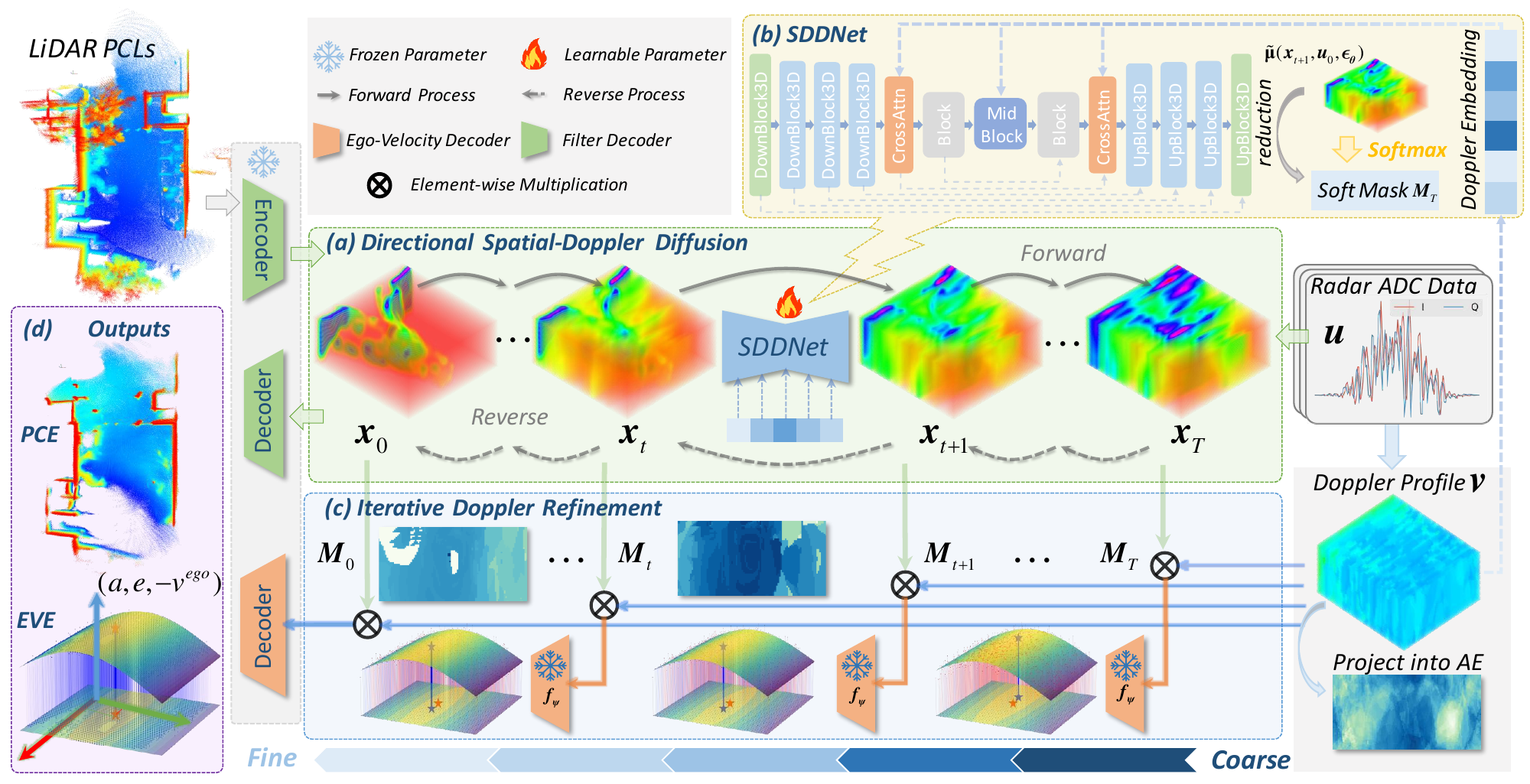}
	\caption{The overview of our SDDiff.}
	\label{fig:overview}
\end{figure*}

\noindent \textbf{Forward process}
Similar to DDPM~\cite{ho2020denoising}, we construct a parameterized Markov chain $\{\bm{x}_0,\bm{x}_1,\cdots,\bm{x}_T\}$ to model the transition between radar and LiDAR distributions as shown in Fig.~\ref{fig:overview}. Given radar's spatial occupancy $\bm{u_0}$ aligned with LiDAR's $\bm{x_0}$, the forward diffusion process is defined as gradually adding Gaussian noise~$\bm{\epsilon }\thicksim \mathcal{N}(0,\mathbf{I})$ to the target representation along the direction toward the radar prior $\bm{u_0}$ as follows:
\begin{align}\label{forward}
q({{\bm{x}}_t}|{{\bm{x}}_{t - 1}},{{\bm{u}}_0}): = {\cal N}({{\bm{x}}_t};{\alpha _t}{{\bm{x}}_{t - 1}} + (1 - {\alpha _t}){{\bm{u}}_0},\lambda _t^2{\bf{I}})
\end{align}
where $\alpha _t$, ${\lambda }_t$ denote pre-defined weight, variance schedule. 

\begin{theorem}\label{th1}
    Given the directional diffusion process with control signal $\bm{u_0}$, sampling the state ${{\bm{x}}_t}$ at an arbitrary timestep $t$ are expressed in closed form:
    \begin{align}\label{forwardone}
q({{\bm{x}}_t}|{{\bm{x}}_0},{{\bm{u}}_0}) := {\cal N}({{\bm{x}}_t};\prod_{k = 1}^t {{\alpha _k}} {{\bm{x}}_0} + (1 - \prod_{k = 1}^t {{\alpha _k}} ){{\bm{u}}_0}, \beta^2_t {\bf{I}})
    \end{align}
    where $\beta^2_t=\sum_{k = 1}^t {{{\bar \alpha _t^2}}\lambda _k^2/{{\bar \alpha _k^2}}}$ and $\bar \alpha _n=\prod_{i = 1}^n {{\alpha _i}}$.
\end{theorem}
The proof of Theorem~\ref{th1} is included in the appendices of the supplementary material. Eqn.~\ref{th1} shows that directional diffusion process starts sampling from the radar prior distribution and diffuses toward the LiDAR SSDR. This facilitates fewer sampling steps and mitigates ambiguous outcomes.

\noindent \textbf{Reverse process}
The key of the reverse process is to estimate the posterior distribution $
p_{\boldsymbol{\theta }}\left( \boldsymbol{x}_{t-1}|\boldsymbol{x}_t,\boldsymbol{u}_0 \right) 
$ with the radar prior $\bm{u}_0$. Following most of the literature on generative models~\cite{rombach2022high}, we aim to choose a tractable distribution $
\mathcal{N}\left( \boldsymbol{x}_{t-1};\boldsymbol{\mu }_{\theta}\left( \boldsymbol{x}_t,\boldsymbol{u}_0 \right) ,\bm{\varSigma }_{\boldsymbol{\theta }}\left( \boldsymbol{x}_t,\boldsymbol{u}_0 \right) \right) 
$  by optimizing the variational bound on negative log likelihood of $p(\bm{x}_0|\bm{u}_0)$.
\begin{align}\label{elbo}
\mathop {\min }\limits_\theta  \sum\limits_{t = 1}^T {\mathbb{E}\left[ {{D_{KL}}\left[ {q\left( {{{\bm{x}}_{t - 1}}|{{\bm{x}}_t},{{\bm{x}}_0},{{\bm{u}}_0}} \right)||{p_\theta }\left( {{{\bm{x}}_{t - 1}}|{{\bm{x}}_t},{{\bm{u}}_0}} \right)} \right]} \right]}
\end{align}
where $D_{KL}[\cdot||\cdot]$ denotes the Kullback-Leibler (KL) divergence.

\begin{theorem}\label{reverse}
For directional diffusion, with the forward process defined in Eqn.~\ref{forwardone}, the posterior distribution of the latent variable is tractable when conditioned on $\bm{x}_0$ and radar's prior $\bm{u}_0$, and is given in explicit form as follows:
\begin{equation}\label{q}
    % \begin{aligned}
p\left( \boldsymbol{x}_{t-1}|\boldsymbol{x}_t,\boldsymbol{x}_0,\boldsymbol{u}_0 \right) \propto \exp \left( -\frac{\left( \boldsymbol{x}_{t-1}-\boldsymbol{ \tilde{\mu}_t  } (\bm{x}_{0,t},\bm{u}_0) \right) ^2}{2\sigma_t ^2} \right) 
% \end{aligned}
\end{equation}
\begin{equation}\label{param1}
    \begin{aligned}
\text{where\ } \boldsymbol{\tilde{\mu}_t }&=\frac{\alpha _t\beta _{t-1}^{2}}{\beta _{t}^{2}}\boldsymbol{x}_t+\frac{\beta _{t}^{2}-\alpha _t\beta _{t-1}^{2}-\lambda _{t}^{2}\bar{\alpha}_{t-1}}{\beta _{t}^{2}}\boldsymbol{u}_0\\
&+{\lambda _{t}^{2}\bar{\alpha}_{t-1}}\boldsymbol{x}_0/{\beta _{t}^{2}} \quad \text{and} \quad \sigma_t ^2=\lambda _{t}^{2}\beta _{t-1}^{2}/\beta _{t}^{2}\\
\end{aligned}
\end{equation}
\end{theorem}
Derivation details on Theorem~\ref{reverse} are provided in the supplementary material appendix due to space limitations.

According to Eqn. (\ref{forwardone})(\ref{q})(\ref{param1}), the mean $
\boldsymbol{\tilde{\mu}}_{\boldsymbol{t}}\left( \boldsymbol{x}_t,\boldsymbol{u}_0 \right) 
$ of the posterior $
p_{\boldsymbol{\theta }}\left( \boldsymbol{x}_{t-1}|\boldsymbol{x}_t,\boldsymbol{u}_0 \right) 
$ conditioned by the previous latent state $\boldsymbol{x}_t$ and the radar prior $\boldsymbol{u}_0$ is further parameterized as follows:
\begin{equation}
\boldsymbol{\tilde{\mu}}_{\boldsymbol{\theta }}\left( \boldsymbol{x}_t,\boldsymbol{u}_0 \right) =\frac{1}{\alpha _t}\boldsymbol{x}_t+\frac{\alpha _t-1}{\alpha _t}\boldsymbol{u}_0+\frac{\lambda _{t}^{2}}{\alpha _t\beta _t}\boldsymbol{\epsilon_{\boldsymbol{\theta }} }
\end{equation}
Therefore, given homogeneous radar prior~$\bm{u}_0$, the objective function in Problem~\ref{elbo} is simplified as:
\begin{equation}
    \mathcal{L} _{\mathrm{Spatial}}=\mathbb{E} _{\boldsymbol{x}_t,\boldsymbol{u}_0}\left[ \frac{\lambda _{t}^{2}}{2\alpha _{t}^{2}\beta _{t-1}^{2}}||\boldsymbol{\epsilon }-\boldsymbol{\epsilon }_{\boldsymbol{\theta }}\left( \boldsymbol{x}_t,\boldsymbol{u}_0 \right) ||_{2}^{2} \right] 
\end{equation}

\subsection{Iterative Doppler Refinement}\label{IDR}
Point cloud density variations across  indoor and outdoor environments and multi-path-induced ghost points hinder model generalization. Drawing from Fig.~\ref{fig:motivation}, we design Iterative Doppler Refinement to mitigate ghost artifacts while progressively enhancing spatial representations for more accurate ego-motion estimation. Specifically, Doppler velocity, \textit{i.e.} relative radial velocity $
v_{i,j}^{r}$, is determined by the radar's ego-velocity $\bm{v}^{ego}$ and the target's azimuth $a_i$ and elevation angle $e_j$ when the target is stationary relative to the ground.
\begin{equation}\label{3dv}
    v_{i,j}^{r}=\left[ \cos a_i\cos e_j,\sin a_i\cos e_j,\sin e_j \right] \left[ v_{x}^{ego},v_{y}^{ego},v_{z}^{ego} \right] ^T
\end{equation}
According to the Cauchy-Schwarz inequality, the peak height of the distribution surface of $v^{r}$ \textit{w.r.t.} $a_i$ and $e_j$ corresponds to the magnitude of the radar's velocity, while the peak position indicates the azimuth and elevation angles of the radar's motion as shown in Fig.~\ref{fig:overview}(d). Considering that static targets dominate the radar field of view in scene-level PCE, we utilize Doppler-consistency as a critic to refine spatial occupancy. Based on Eqn.~\ref{3dv} we train a differentiable model $f_{\psi}\left( \boldsymbol{v} \right) $
 for velocity regulation. Given a progressively refined spatial representation $\bm{x}_t$, a reduction operation along range followed by softmax is applied to obtain a soft mask $\bm{M}_t$. Further, we define a Doppler-consistency loss to jointly refine spatial occupancy and ego velocity.
 \begin{equation}\label{loss_doppler}
\mathcal{L} _{\mathrm{Doppler}}=\underset{\boldsymbol{M}_t,\boldsymbol{v}_t}{\mathbb{E}}\left[ \frac{\lambda _{t}^{2}\bar{\alpha}_{t-1}^{2}}{2\beta _{t}^{2}\beta _{t-1}^{2}}||\boldsymbol{v}^{ego}-f_{\psi}\left( \boldsymbol{M}_t\odot \boldsymbol{v} \right) ||_{2}^{2} \right] 
 \end{equation}
where $\odot$ donates Hadamard product, the weights are determined by Eqn.~\ref{param1}.
\subsection{Overview of Our SDDiff}\label{OOS}
Fig.~\ref{fig:overview} presents the overview of our SDDiff. During training, the LiDAR representations are softened with a Gaussian kernel to transition to a probabilistic spatial occupancy $\bm{x}_0$. Subsequently, they and radar priors $\bm{u}_0$ are fed into the directional spatial-Doppler diffusion module. We adopt SDDNet as the denoising model, designed on a standard 3D U-Net architecture with paird up-down blocks incorporating ResNet and attention mechanism. To optimize computational efficiency, radar Doppler profiles $\bm{v}$ are embedded \textit{w.r.t.} chirp channels before injected into each 3D U-Net layer for cross-attention. Subsequently, Iterative Doppler Refinement with the latent variable $\bm{x}_t$ and Doppler profile $\bm{v}$ as inputs, is initialized to mitigate density variations and ghosting effects. The overall training loss for SDDNet is formulated as:
\begin{gather}
    \mathcal{L} _{\mathrm{SDDNet}}= \mathcal{L} _{\mathrm{Spatial}} + \omega \mathcal{L} _{\mathrm{Doppler}}
\end{gather}
where $\omega$ denotes the weight balancing spatial and Doppler losses. Notably, only the parameters of SDDNet are optimized during training. The inference process resembles Stable Diffusion, iteratively sampling the SDDR through Markov chains. The refined SDDR is subsequently employed for ego velocity estimation.
\section{Experiments Setup}
\subsection{Dataset}
We evaluate the proposed method using both the publicly available Coloradar dataset and a self-collected dataset across different indoor and outdoor scenarios.

\noindent {\textbf{ColoRadar Dataset.}} We conduct our method on ColoRadar Dataset~\cite{kramer2022coloradar}, which includes raw ADC sample data for single-chip radar, dense LiDAR point clouds, and odometry pose information. We use LiDAR point clouds as the ground truth for PCE and velocity solved through odometry for EVE. The ColoRadar contains both indoor and outdoor scenes with a total of 52 sequences. For a fair comparison with other learning-based baselines, we select the same 36 sequences as the training set and others for testing.

\begin{figure}[h]
  \centering
   \includegraphics[width=.8\linewidth]{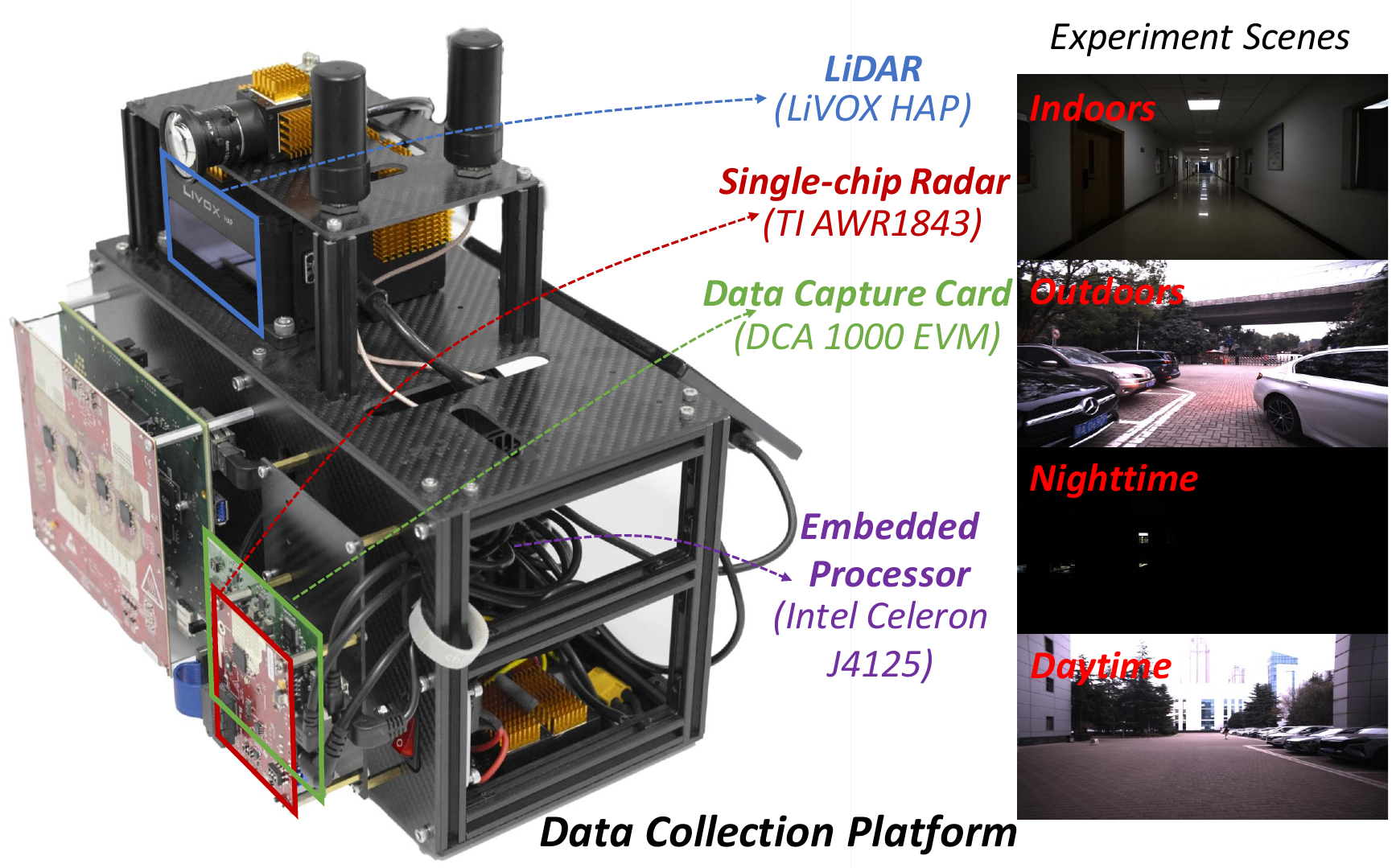}
   \caption{Our customized hand-held data collection platform.}
   \label{fig:setup}
\end{figure}

\noindent {\textbf{Self-Collected Dataset.}} 
We collected real-world data from diverse environments using the platform shown in Fig.~\ref{fig:setup} to assess the model's generalization. The dataset comprises 10,371 frames, with 10\% used for fine-tuning and 90\% for testing. Reliable odometry for EVE ground truth was obtained using Fast-Livo~\cite{zheng2022fast}.

\subsection{Implementation Details}
We implement our SDDiff using Pytorch 1.11.0 with CUDA 12.4. The parameters $\omega$ of the weighted spatial and Doppler loss are set to 0.01. We apply Gaussian filtering with sigma values of [0.2, 0.5, 1] as the filter encoder and perform offline training for the associated decoder. The Ego-Velocity Decoder is trained according to Eqn.~\ref{3dv}, with the Azimuth-Elevation~(AE) Doppler profile as input. The forward process variances are set to constants increasing linearly from $\bar \alpha_1=0.01$ to $\bar \alpha_T=0.99$ and the noise scale $\lambda_k$ is set to 0.1. To illustrate the effectiveness of the directional diffusion strategy, we set the sampling step size $T=20$. We train SDDNet for 100 epochs on ColoRadar Dataset with AdamW optimizer and a learning rate $10^{-4}$. It takes about 5 days to train our model with a machine using three NVIDIA GeForce RTX 4090 GPUs and Intel Xeon Gold 6226R CPU.

\begin{figure*}[t]
  \centering
   \includegraphics[width=.85\linewidth]{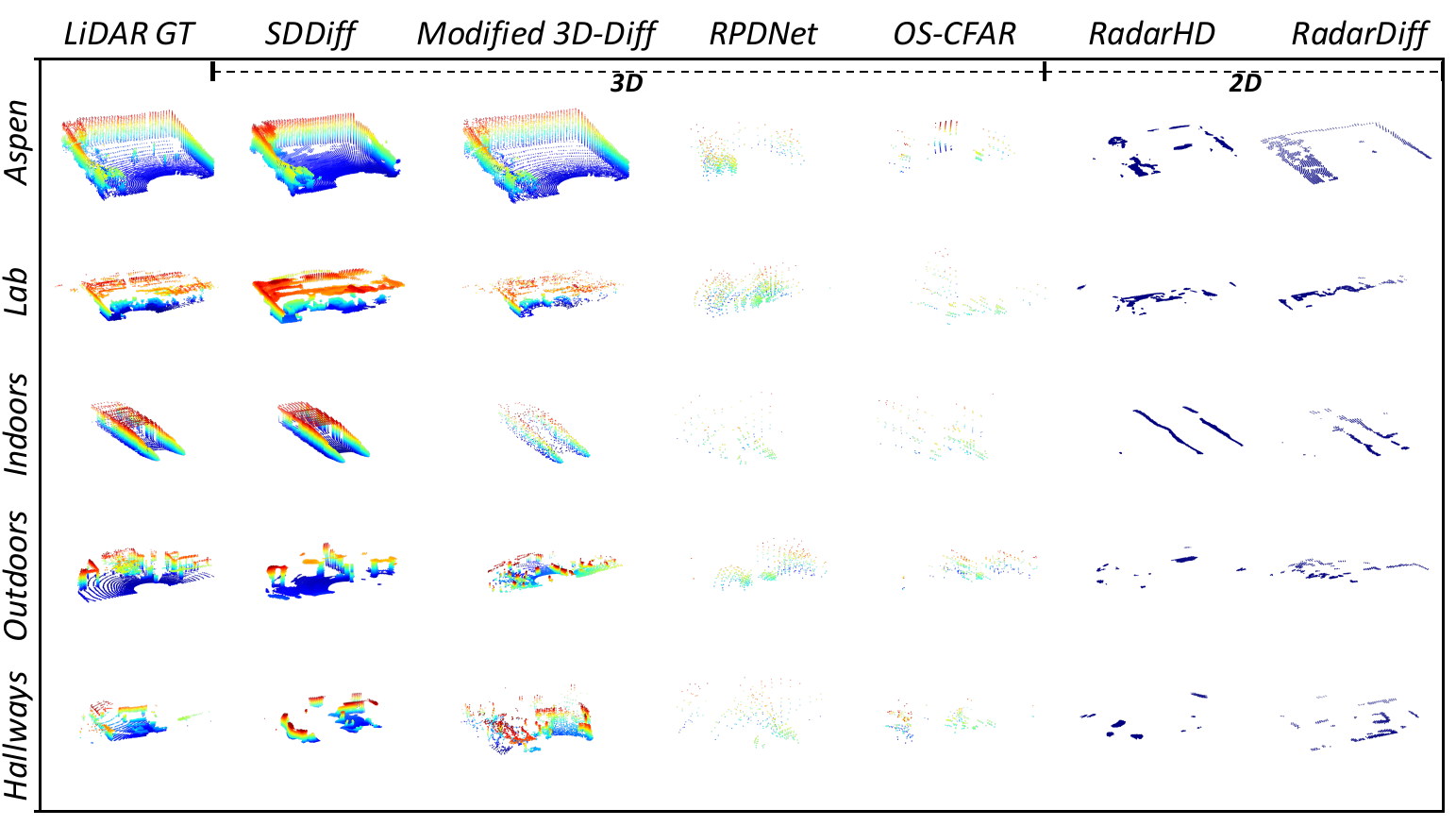}
   \caption{Qualitative results of different methods and ours on the ColoRadar dataset.}
   \label{fig:showcase}
\end{figure*}

\subsection{Evaluation Metric}
\noindent \textbf{Baseline.} We evaluate the PCE performance of SDDiff compared to traditional method {OS-CFAR}~\shortcite{blake1988cfar}, as well as generative methods {RPDNet}~\shortcite{cheng2022novel}, {RadarHD}~\shortcite{prabhakara2023high} and {RadarDiff}~\shortcite{10592769}. Since no generative models exist for scene-level 3D PCE, we modify the naive diffusion model with radar spatial occupancy as input for comparison with our approach, denoted as modified 3D-Diff. For EVE performance evaluation, we compare our method with ICP~\shortcite{besl1992method}, RANSAC~\shortcite{kellner2013instantaneous}, and RadarEVE~\shortcite{pang2024radarmoseve}.

 \textbf{PCE Metrics.} We use Earth Mover's Distance~(EMD) and Chamfer Distance~(CD) to quantify the similarity between the extracted point clouds $\mathcal{P}$ and the ground truth~$\mathcal{Q}$. Additionally, we adopt the definition of clutter points as provided in RPDNet~\cite{cheng2022novel}, as follows:
\begin{align}
    \mathcal{P}_{\text{clutter}}=\left\{ \boldsymbol{p}\in \mathcal{P},\ s.t,\ d\left( \boldsymbol{q,p} \right) >\tau _1,\ \forall \boldsymbol{q}\in \mathcal{Q} \right\} 
\end{align}
where $d\left( \boldsymbol{q,p} \right)$ denotes the Euclidean Distance between $\bm{p}$ and $\bm{q}$. Similarly, we define the `shot' LiDAR points as those generated points that can effectively represent the scene.
\begin{align}
    \mathcal{Q}_{\text{shot}}=\left\{ \boldsymbol{q}\in \mathcal{Q},\,\,\exists \boldsymbol{p}\in \mathcal{P},\ s.t,\,\,d\left( \boldsymbol{p,q} \right) < \tau _2 \right\} 
\end{align}
We define scene representation level and valid point ratio for point cloud generation quality assessment, and effective generation density for density evaluation.
\begin{itemize}
    \item {\textit{Valid Point Ratio~(VPR)}} is defined as $1-|\mathcal{P}_{\text{clutter}}|/|\mathcal{P}|$, reflecting the reliability of the generated point clouds.
    \item {\textit{Scene Representation Level~(SRL)}} is defined as $|\mathcal{Q}_{\text{shot}}|/|\mathcal{Q}|$, quantifying the effectiveness of the point cloud generation.
    \item {\textit{Effective Generation Density~(EGD)}} is characterized as $|(\mathcal{P} - \mathcal{P}_{\text{clutter}}|)/|\mathcal{Q}_{\text{shot}}|$, referring to the density of the point clouds.
\end{itemize}

 \textbf{EVE Metrics.} We evaluate EVE using Mean Absolute Error~(MAE) and cumulative velocity error density in both indoor and outdoor scenarios.
\section{Results and Analysis}

\subsection{Point Cloud Extraction Results}
Qualitatively, we showcase the PCE results of various approaches, as shown in Fig.~\ref{fig:showcase}. Our method demonstrates superior density and accuracy in PCE compared to alternatives. Quantitatively, we compare the point clouds generated by different schemes on the Coloradar Dataset against the LiDAR ground truth, using metrics EMD and CD. As presented in Tab.~\ref{tab:edm_dataset}, points generated by our method exhibit the closest alignment with the LiDAR ground truth. We also evaluate our method on the self-collected dataset to demonstrate its generalization. As shown in Tab.~\ref{tab:edm_self}, our approach consistently outperforms others when applied to new scenarios.

\begin{figure}[htbp]
  \centering
   \includegraphics[width=.91\linewidth]{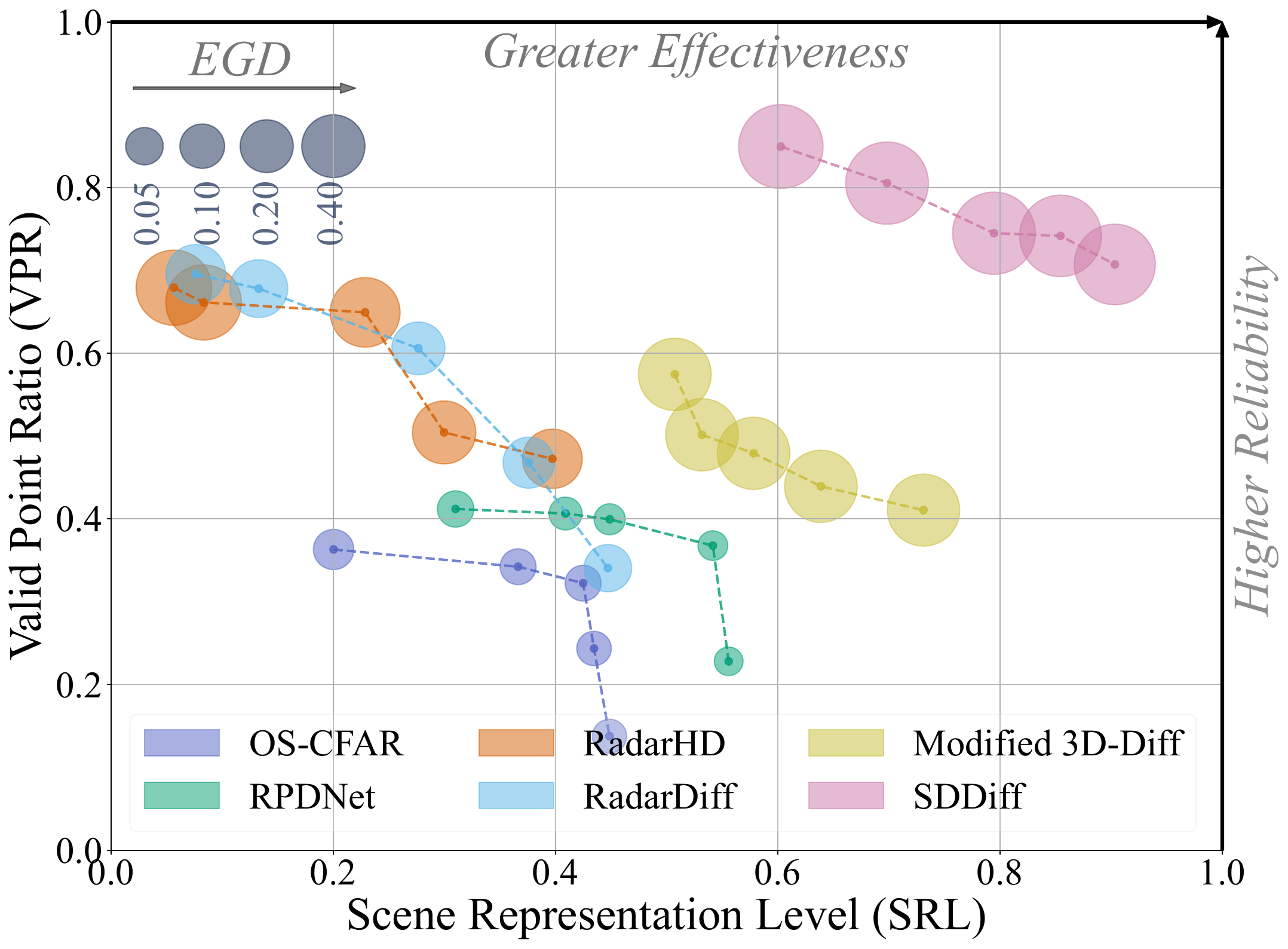}
   \caption{The evaluation results for effectiveness, reliability, and density of PCE. A larger scatter radius corresponds to a higher EGD.}
   \label{fig:tradeoff}
\end{figure}

\textbf{Reliability and Effectiveness Analysis}: Different models show variations in the number of points and clutter at different thresholds. To comprehensively assess the reliability and effectiveness of the generated point clouds, we evaluate the clutter rate, representation level, and density of various PCE methods. As illustrated in Fig.~\ref{fig:tradeoff}, 2D methods exhibit high VPR and low SRL due to the lack of elevation information. our method achieves a balanced trade-off between effectiveness and reliability, outperforming the state-of-the-art baseline with a 30\% improvement in VPR and a 33\% increase in SRL. Moreover, our method ensures the highest valid point cloud density compared to other approaches. This is attributed to the reciprocal benefit mentioned in Fig.~\ref{fig:motivation}, which enhances the model's robustness to density variations and ghosting effects.

\begin{table*}[h]
    \centering
    \fontsize{8}{8}\selectfont
    \begin{tabular}{cc@{\hskip 6pt}cc@{\hskip 6pt}cc@{\hskip 6pt}cc@{\hskip 6pt}cc@{\hskip 6pt}cc@{\hskip 6pt}cc@{\hskip 6pt}c}
        \toprule[0.5pt]
        \toprule
        \multirow{2}{*}{ \makecell{Object Dense   \\ Detection Methods} }&
        \multicolumn{2}{c}{Classroom}& \multicolumn{2}{c}{Armyroom} &\multicolumn{2}{c}{Hallways}&\multicolumn{2}{c}{Labroom}&\multicolumn{2}{c}{Aspen\_room}
        &\multicolumn{2}{c}{Outdoors}&\multicolumn{2}{c}{Longboard} \cr
        \cmidrule(lr){2-3}
        \cmidrule(lr){4-5}
        \cmidrule(lr){6-7}
        \cmidrule(lr){8-9}
        \cmidrule(lr){10-11}
        \cmidrule(lr){12-13}
        \cmidrule(lr){14-15}
        & EMD $\downarrow $ & CD $\downarrow $& EMD $\downarrow $& CD $\downarrow $& EMD $\downarrow $& CD $\downarrow $& EMD & CD $\downarrow $& EMD $\downarrow $& CD $\downarrow $& EMD $\downarrow $& CD $\downarrow $& EMD $\downarrow $& CD$\downarrow $\cr
        \cmidrule(lr){1-15}

        {OS-CFAR}~\shortcite{blake1988cfar} & 1.03 & 1.09 & 1.06 & 1.05 & 1.15 & 1.18 & 1.07 & 1.29 & 1.19 & 1.28 & 1.54 & 1.37  & 4.26 & 3.57\cr
        {RPDNet}~\shortcite{cheng2022novel} & 0.97 & 0.85 & 1.31 & 1.07 & 1.40 & 1.10 & 1.09 & 1.01 & 1.30 & 1.08 & 1.69 & 1.24 & 2.29 & 1.67\cr
        {RadarHD}$^\dag$~\shortcite{prabhakara2023high} & 0.57 & 0.53 & 0.77 & 0.76 & 1.63 & 1.29 & 1.40 & 1.11 & 0.85 & 0.77 & 3.20 & 2.28 & 4.55 & 3.95 \cr
        {RadarDiff}$^\dag$~\shortcite{10592769} & 0.72 & 0.54 & 0.87 & 0.68 & 1.46 & 1.03 & 1.13 & 0.83 & 1.32 & 0.89 & 2.14 & 1.45 & 4.16 & 2.73\cr
        {Modified 3D-Diff} & 0.58 & 0.56 & 0.77 & 0.77 & 0.85 & 0.84 & 0.90 & 0.89 & 0.88 & 0.88 & 1.32 & 1.29 & 1.20 & 1.12 \cr
        \rowcolor{gray50!20} \textbf{SDDiff (Ours)} &  {\textbf{0.25}} &  {\textbf{0.24}} &  {\textbf{0.31}} &  {\textbf{0.31 }}&  {\textbf{0.56}} &  {\textbf{0.64}}&  {\textbf{0.47}} &  {\textbf{0.53}} &  {\textbf{0.38}} &  {\textbf{0.38}}& {\textbf{0.75}}& {\textbf{0.80}}  &  {\textbf{0.81}} &  {\textbf{0.88}}\cr
        \bottomrule
        \bottomrule[0.5pt]
    \end{tabular}\vspace{0cm}
    \caption{Point Cloud Extraction results on the Coloradar Dataset test split. $^\dag$ indicates 2D methods for distinction.}
    \label{tab:edm_dataset}
\end{table*}

\begin{table*}[h]
    \centering
    \fontsize{8}{8}\selectfont
    \begin{tabular}{ccccccccccccc}
        \toprule[0.5pt]
        \toprule
        \multirow{3}{*}{ \makecell{Object Dense   \\ Detection Methods} }&
        \multicolumn{6}{c}{Indoors}& \multicolumn{6}{c}{Outdoors} \cr
        \cmidrule(lr){2-7}
        \cmidrule(lr){8-13}
        & \makecell{EMD$\downarrow $ \\ (50\%)} & \makecell{CD$\downarrow $ \\ (50\%)} & \makecell{MHD$\downarrow $ \\ (50\%)} & \makecell{EMD$\downarrow $ \\ (90\%)} & \makecell{CD$\downarrow $ \\ (90\%)} & \makecell{MHD$\downarrow $ \\ (90\%)} & \makecell{EMD$\downarrow $ \\ (50\%)} & \makecell{CD$\downarrow $ \\ (50\%)} & \makecell{MHD$\downarrow $ \\ (50\%)}& \makecell{EMD$\downarrow $ \\ (90\%)} & \makecell{CD$\downarrow $ \\ (90\%)} & \makecell{MHD$\downarrow $ \\ (90\%)} \cr
        \cmidrule(lr){1-13}

         {OS-CFAR}~\shortcite{blake1988cfar} & 0.78 & 1.01 & 0.88 & 1.23 & 1.62 & 1.54 & 2.32 & 2.21 & 2.36 & 4.72 & 4.57 & 5.51 \cr
         {RPDNet}~\shortcite{cheng2022novel} & 0.96 & 0.91 & 0.83 & 1.46 & 1.53 & 1.34 & 1.51 & 1.39 & 1.30 & 2.63 & 2.01 & 2.49 \cr
         {RadarHD$^\dag$}~\shortcite{prabhakara2023high} & 0.55 & 0.51 & 0.36 & 1.03 & 0.89 & 0.64 & 1.91 & 1.53 & 1.57 & 4.43 & 3.24 & 5.32 \cr
         {RadarDiff$^\dag$}~\shortcite{10592769} & 0.54 & 0.77 & 0.31 & 1.22 & 2.01 & 1.56 & 1.63 & 1.68 & 1.68 & 3.35&3.25&3.33 \cr
         {Modified 3D-Diff} & 0.42 & 0.41 & 0.33 & 1.14 & 0.98 & 0.81 & 0.79 & 0.76 & 0.64 & 2.05&1.16&1.86 \cr
        \rowcolor{gray50!20} \textbf{SDDiff (Ours)}  & \textbf{0.33} & \textbf{0.33} & \textbf{0.21} & \textbf{0.79} & {\textbf{0.79}} & \textbf{0.44}& \textbf{0.62} & \textbf{0.61} & \textbf{0.50} & \textbf{0.85}&\textbf{0.84}&\textbf{0.75} \cr
        \bottomrule
        \bottomrule[0.5pt]
    \end{tabular}\vspace{0cm}
    \caption{Point Cloud Similarity Comparison at different percentiles on datasets collected from real-world scenarios.}
    \label{tab:edm_self}
\end{table*}

\subsection{Ego Velocity Estimation Results}
As shown in Tab.~\ref{tab:eve}, our method improves EVE by 30\% and 59\% over the state-of-the-art in indoor and outdoor scenarios, respectively. This is attributed to our PCE process, which provides more accurate and reliable point clouds for EVE. In contrast, other methods degrade outdoors due to relying on the sparse points processed by the data capture card.

\begin{table}[h]
    \centering
    \fontsize{7}{8}\selectfont
    \begin{tabular}{c@{\hskip 4pt}c@{\hskip 5pt}c@{\hskip 4pt}c@{\hskip 4pt}c@{\hskip 4pt}c@{\hskip 4pt}c}
        \toprule[0.5pt]
        \toprule
        \multirow{2}{*}{\makecell{EVE\\ Methods} }&
        \multirow{2}{*}{\makecell{MAE$\downarrow $\\ ($\cdot$ m/s)} }&
        \multicolumn{5}{c}{Cumulative Density of Error/\%~(Indoors/Outdoors)~$\uparrow$} \cr
        \cmidrule(lr){3-7}
        & & $\le$0.05m/s & $\le$0.1m/s & $\le$0.15m/s & $\le$0.2m/s & $\le$0.25m/s \cr
        \cmidrule(lr){1-7}
        {ICP} & 0.60/0.77 & 5/6 & 12/10 & 17/15 & 23/20 & 29/25  \cr
         {RANSAC} & 0.31/0.57 &  {\textbf{35}}/13 & 55/27 & 63/37 & 66/44 & 67/47  \cr
        {RadarEVE}  & 0.13/0.27 & 29/8 & 51/14 & 66/22 & 79/36 & 87/50 \cr
        \rowcolor{gray50!20} \textbf{SDDiff} &  {\textbf{0.09}}/{\textbf{0.11}} & 31/{\textbf{29}} &  {\textbf{59}}/{\textbf{55}} &  {\textbf{78}}/{\textbf{74}}&  {\textbf{89}}/{\textbf{86}} &  {\textbf{95}}/{\textbf{93}}\cr
        \bottomrule
        \bottomrule[0.5pt]
    \end{tabular}\vspace{0cm}
    \caption{The EVE performance results}
    \label{tab:eve}
\end{table}

\subsection{Ablation Study}
We conduct an ablation study to validate the gains introduced by PCE module \textit{i.e.} Directional Spatial-Doppler Diffusion and EVE module \textit{i.e.} Iterative Doppler Refinement. As shown in Tab.~\ref{tab:ablation}, the PCE without Doppler information incurs a 15\% reduction in PCE performance compared to the whole SDDiff. Nevertheless, it still outperforms Modified 3D-Diff, which suffers from ambiguous outcomes due to its diffusion scheme. Moreover, it speeds up inference by 3.13$\times$ compared to Modified 3D-Diff, owing to the more efficient sampling prior distribution. The EVE without reliable spatial occupancy leads to a 1.2$\times$ and 3.4$\times$ increase in indoor and outdoor errors, respectively. This demonstrates that SDDiff effectively harnesses the potential gains from the synergy between PCE and EVE.
\begin{table}[h]
    \centering
    \fontsize{8}{8}\selectfont
    \begin{tabular}{ccccccc}
        \toprule[0.5pt]
        \toprule
        \makecell{PCE\\Module} &
        \makecell{EVE\\Module} &
        \makecell{VPR\\(\%)$\uparrow$} &
        \makecell{SRL\\(\%)$\uparrow$} &
        \makecell{EGD \\ ($\times$)$\uparrow$}  &
        \makecell{V.E.\\(m/s)$\downarrow$ }&
        \makecell{Speed \\ ($\times$)$\uparrow$} \cr
        \cmidrule(lr){1-7}
        
        \CheckmarkBold & \XSolidBrush & 65.5 & 69.2 & 0.98 & - & 3.13 \cr
        \XSolidBrush & \CheckmarkBold & - & - & - & 0.11/0.37 & -  \cr
        \CheckmarkBold & \CheckmarkBold & 77.1 & 79.6 & 1.17 & 0.09/0.11 & 2.57 \cr
        \bottomrule
        \bottomrule[0.5pt]
    \end{tabular}\vspace{0cm}
        \caption{Ablation quantitative results on EVE and PCE.}
    \label{tab:ablation}
\end{table}

\section{Conclusion}
This paper introduces SDDiff, a directional Spatial-Doppler diffusion model to simultaneously enable dense point cloud extraction and accurate ego velocity estimation. To reduce sampling wastage and mitigate ambiguous outcomes, we design a directional diffusion with radar priors to streamline the sampling. Additionally, we design Iterative Doppler Refinement to defend against density variations and ghosting effects. The experiments demonstrate that SDDiff enhances EVE accuracy while improving both the effectiveness and reliability of PCE.

\appendix
\bibliographystyle{named}
\bibliography{ijcai25}

\end{document}